\documentclass{article}

\author{
  Rushikesh Zawar \\
   \\
  \texttt{}
  \And
  Vladimir Kim \\
   \\
  \texttt{
}
}
\PassOptionsToPackage{numbers}{natbib}
\usepackage[final,dblblindworkshop]{neurips_2026}
\workshoptitle{Physical World AI: Geometry, Characteristics, and Multimodal Sensing}

\usepackage{graphicx}
\usepackage{booktabs}
\usepackage{array}
\usepackage{multirow}
\usepackage{amsmath,amssymb}
\usepackage{microtype}
\usepackage{hyperref}
\hypersetup{hidelinks}

\newcommand{\good}{\mathrm{good}}

\begin{document}

\title{Is This 3D Asset Usable? Beyond the Beauty Render for Asset Curation}

\maketitle

\begin{abstract}
Large 3D repositories make it easy to collect assets, but difficult to determine
whether they can be reused reliably across scene assembly, editing, rendering,
game production, and training-data construction. An asset can look plausible
in RGB while hiding disconnected geometry, problematic UV organization, or
materials that fail to resolve correctly after import. We introduce \emph{Diagnostic Rendered Evidence (DRE)}: aligned beauty, wireframe,
and UV-seam-plus-wireframe views scored by pass-specific frozen ViT
classifiers and combined through late fusion. On 1,234
human labeled TexVerse assets, nested grouped evaluation achieves 86.36\%
balanced accuracy and 88.32\% bad-asset recall. On a separate, balanced 764 external reference test, DRE achieved 88.9\% balanced
reference-label accuracy. Our method requires only an imported asset and renderer-derived evidence, without
repository identity, creator, popularity, text, or source metadata at
inference. Our results support cross-corpus
portability and show that purpose built diagnostic views expose defects that just the beauty views can miss. Webpage: https://rbz-99.github.io/is-this-3d-asset-usable/

\end{abstract}

\section{Introduction}
\label{sec:introduction}

Consider a common workflow where we download a model from a large 3D repository. It can look perfectly fine and still be difficult to use. A few RGB views look convincing: the shape is recognizable, the texture appears detailed, and nothing is obviously broken. Yet once the asset enters a 3D workflow, different problems can emerge. The mesh may contain disconnected or irregular geometry, the UV layout may complicate texturing, or materials may fail to resolve correctly after import. None of these problems necessarily prevents the asset from producing an attractive preview.

Repository assets are increasingly reused as structured 3D objects: retrieved and arranged into scenes \cite{yang2024holodeck,aguinakang2024openuniverse,rahamim2024layascene}, edited \cite{erkoc2024preditor3d}, rigged for animation \cite{liu2025riganything}, or used in modern image-to-3D pipelines that explicitly represent geometry, UVs, textures, and PBR attributes \cite{boss2024sf3d,siddiqui2024assetgen,xiang2026trellis2}. Across these uses, plausible RGB appearance does not guarantee reliable reuse.

Large repositories such as ShapeNet \cite{shapenet2015}, Objaverse and Objaverse-XL \cite{deitke2023objaverse,deitke2023objaversexl}, and TexVerse \cite{zhang2025texverse} make this a repository-scale curation problem:
\emph{How can we determine whether a 3D asset is fit for reuse before selecting it from a large repository?}

This requires expert review, that considers appearance, geometry, UV structure, and material behavior. We therefore construct a 1,234-asset TexVerse set annotated by 3D researchers and introduce \textbf{diagnostic rendered evidence}: aligned beauty, wireframe, and UV-seam-plus-wireframe views that expose complementary properties of the same asset. Separate models score each evidence type before late fusion produces the final keep/reject decision. Our scope is \emph{operational fitness} for reusable textured/PBR assets rather than universal aesthetic quality or asset provenance. On our human-labeled assets, our method reaches 86.36\% balanced accuracy and 88.32\% bad-asset recall. Controlled ablations show that the structural diagnostic passes provide information beyond a matched beauty-only representation. Our contributions are: 1: We introduce \emph{diagnostic rendered evidence} for 3D asset
    curation using aligned beauty, wireframe, and UV-seam-plus-wireframe views. 2: We develop a robust rendering pipeline for heterogeneous repository
    assets, including guarded material fallbacks and explicit
    material-resolution provenance. 3: We provide a 1,234-asset expert-annotated benchmark for reusable
    textured/PBR assets, with a keep/reject rubric covering geometry, UV,
    material, render, and composition failures.

\section{Related Work and Scope}
\label{sec:related}

\textbf{3D assets as reusable components:}
Recent systems increasingly treat existing 3D assets as reusable components.
Holodeck retrieves and arranges Objaverse assets for scene generation
\cite{yang2024holodeck}; Open-Universe similarly retrieves meshes from
uncurated databases and optimizes their placement
\cite{aguinakang2024openuniverse}; and Lay-A-Scene studies object arrangement
\cite{rahamim2024layascene}. Existing assets are also edited
\cite{erkoc2024preditor3d} and automatically rigged for animation
\cite{liu2025riganything}. These settings motivate evaluating asset properties
that are not fully captured by a beauty render.

\textbf{3D curation and quality assessment:}
Objaverse++ and MeshFleet use filtering and human review to construct
higher-quality 3D corpora \cite{lin2025objaversepp,boborzi2025meshfleet},
while generative-3D benchmarks evaluate geometry, texture, and cross-view
consistency \cite{he2023t3bench,su2024gt23d,zhang2024mate3d,ebert2025arena}.
Perceptual mesh-quality studies likewise show that geometry and texture jointly
affect human judgments \cite{guo2021textured,nehme2023texturedmesh}.
Full-reference metrics assume a known reference
\cite{nehme2023texturedmesh,yang2025fmqm}; repository curation instead requires
judging a single asset without one.

\textbf{Structured evidence for asset quality:}
Modern 3D systems explicitly represent geometry, UVs, and materials.
Hunyuan3D separates mesh and texture generation \cite{zhao2025hunyuan3d},
SF3D predicts UV-unwrapped geometry and material properties
\cite{boss2024sf3d}, and AssetGen reconstructs geometry and PBR appearance
\cite{siddiqui2024assetgen}. Related work further models UV and material
structure explicitly \cite{chen2025meshgen,wei2025pbr3dgen,xiang2026trellis2},
while UV-specific methods motivate interpreting parameterization in surface
context \cite{wang2025partuv,richardson2023texture}. Our method uses rendered
views of these structures as no-reference evidence for asset curation.
\section{Methodology}
\label{sec:method}

\textbf{Evidence Generation and Provenance:}
For each imported GLB, we render synchronized diagnostic passes using fixed
camera poses, lighting, background, and framing. Four aligned views are
generated per pass and scored independently before late fusion. The \emph{beauty} pass captures overall appearance and visible rendering or
material failures, but may hide structural defects through texture, shading,
and lighting. The \emph{wireframe} pass exposes mesh structure, including
disconnected components, irregular triangulation, excessive density, and
geometric clutter. These patterns are treated as evidence rather than fixed
quality rules. The \emph{UV-seam-plus-wireframe} pass overlays UV boundaries
on the mesh, allowing seam organization to be interpreted in its 3D surface
context rather than from a flattened atlas alone.

\begin{figure}[t]
\centering
\includegraphics[width=1\columnwidth]{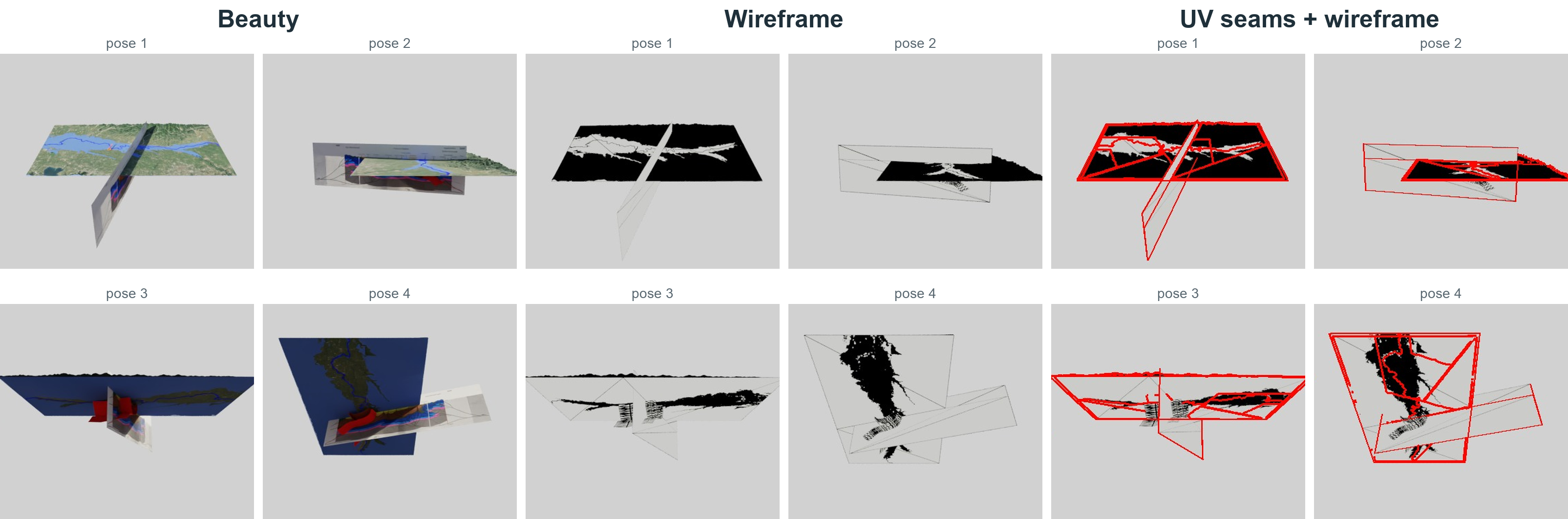}
\caption{Aligned beauty, wireframe, and UV-seam-plus-wireframe evidence. The external baseline assigns this asset \emph{Superior}, but the diagnostic views expose its disconnected planar fragments.}
\label{fig:render-evidence}
\end{figure}

\textbf{Repository-Agnostic Evidence:} DRE is designed around information that
can be obtained from an imported asset rather than from a particular repository.
The renderer uses no asset ID, creator, source label, caption, popularity
statistic, or repository-specific metadata. It first resolves standard
Principled-BSDF base-color inputs, then follows supported graph-connected image
textures and guarded fallbacks; when resolution is not possible, it records an
explicit unresolved-material state rather than inventing a material
interpretation. We also generated 2D UV-island layouts and measurements such as coverage,
angular distortion, and texel density. These signals were reviewed by 3D
experts and evaluated in additional classifier configurations, but provided
little improvement over the surface-aligned diagnostic views. We therefore
retain them for review and audit rather than classifier input, particularly
since UV packing conventions vary across asset categories and authoring
workflows. Figure~\ref{fig:render-evidence} illustrates the synchronized evidence available to both the reviewer and the classifier. No individual diagnostic pass is treated as ground truth. Human review separates several kinds of concern: low overall curation value, geometry integrity or clutter, UV organization, material-resolution anomalies, missing evidence, and other import or composition problems.

\begin{table*}[t]
\centering
\begin{minipage}[t]{0.30\textwidth}
\centering
\scriptsize
\textbf{(a) Selected fusion}\\[-2pt]
\begin{tabular}{lr}
\toprule
Metric & Value \\
\midrule
Accuracy & 85.58\% \\
Balanced accuracy & 86.36\% \\
Bad-asset recall & 88.32\% \\
Good-asset recall & 84.41\% \\
Good-asset precision & 94.44\% \\
Bad predicted good & 43 \\
Good predicted bad & 135 \\
\bottomrule
\end{tabular}

\end{minipage}\hfill
\begin{minipage}[t]{0.65\textwidth}
\centering
\scriptsize
\textbf{(b) Representation-controlled evidence ablation}\\[-2pt]
\begin{tabular}{lrrr}
\toprule
Evidence configuration & Acc. & BA & BR \\
\midrule
Beauty & 73.26 & 71.26 & 66.30 \\
Wireframe & 81.20 & 80.35 & 78.26 \\
UV seams + wire & \textbf{84.60} & \textbf{84.97} & 85.87 \\
Beauty + wireframe & 81.28 & 80.65 & 79.08 \\
Beauty + UV seams + wire & 84.28 & 84.03 & 83.42 \\
Wireframe + UV seams + wire & 84.04 & 84.88 & 86.96 \\
All three passes & 82.90 & 84.15 & \textbf{87.23} \\
\bottomrule
\end{tabular}

\end{minipage}
\caption{Grouped development results on 1,234 human-labeled assets. (a) Selected nested grouped fusion. (b) Ablations holding labels, groups, ViT representation, scoring family, and threshold-selection protocol fixed. Acc.: accuracy, BA: balanced accuracy, BR: bad-asset recall.}
\label{tab:results}
\end{table*}

\textbf{Multi-view Scoring and Fusion:} For pass $m$ and view $v$, a frozen ImageNet ViT-B/16 encoder \cite{dosovitskiy2021vit} produces $z_{m,v}\in\mathbb{R}^{768}$. We summarize the four aligned views with their feature-wise mean and standard deviation. A standardized, class-balanced logistic-regression model is then trained separately for beauty, wireframe, and UV-seam-plus-wireframe evidence. Each model produces a probability $p_m(\good)$ describing how strongly that particular type of evidence supports keeping the asset.

\section{Learning and Evaluation}
\label{sec:evaluation}

We use TexVerse \cite{zhang2025texverse} to create our set
under CC BY, CC BY-SA, or CC0 licenses. Our supervised set contains 1,234 assets annotated by expert 3D researchers:
866 good and 368 bad. An additional 6,685 assets based on the deterministic render QA and their metadata are used as less weighted auxiliary
training data. We use five-fold nested stratified group cross-validation, grouping by creator
or source where available. Inner folds train the pass-specific classifiers,
generate out-of-fold predictions for fusion, and select weak-label weights,
regularization, pass combinations, and the decision threshold. Auxiliary
samples sharing a group with the held-out fold are excluded. Metadata and strict full-PBR inventory provide weak-label
sources, while terms such as \emph{Rodin}, \emph{photogrammetry}, or
\emph{scan} identify low-weight negative candidates rather than evaluation
labels. Face count, texture resolution, creator, and source are retained only
for audit and stratification. Table~\ref{tab:results} reports outer-fold
predictions.

\textbf{Modality Ablation:}
Table~\ref{tab:results} keeps the labels, grouped folds, ViT representation,
classifier family, fusion method, and threshold-selection protocol fixed while
changing only the render evidence. UV-seam-wire achieves the highest balanced
accuracy, while the full three-pass model gives the highest bad recall and
fewest false accepts. Thus, the benefit is not simply due to using more views:
the structural passes provide information complementary to beauty renders. We exclude material-only views because their appearance is difficult to
separate from renderer behavior. A gray render, for example, may indicate an
intentional material, scalar input, vertex color, missing texture, or failed
graph resolution. Material behavior is therefore audited separately rather
than used as a learned visual modality.

\textbf{Baselines:}
The matched beauty-only ViT in Table~\ref{tab:results} is our primary
representation baseline. It keeps the labels, groups, backbone, classifier,
fusion, and threshold-selection protocol fixed, isolating the contribution of
diagnostic evidence. We also evaluate the released Objaverse++ 
\cite{lin2025objaversepp} on our dataset using its
40-view protocol.  Our method uses no
popularity, creator, source, or text features at inference and requires 12
diagnostic images per asset, compared with 40 RGB views for Objaverse++.
Objaverse++ outputs are mapped as High/Superior = keep and Low/Medium =
reject.

\begin{table}[t]
\centering
\caption{Objaverse++ versus DRE (ours) on expert-labeled TexVerse assets and a cross-source reference set.}\label{tab:objaversepp}
\scriptsize
\begin{tabular}{llrrrr}
\toprule
Set & Model & Acc./BA $\uparrow$ & G-Rec. $\uparrow$ & B-Rec. $\uparrow$ & BAR $\downarrow$ \\
\midrule
TexVerse & Objaverse++ & 44.65 / 55.72 & 28.29 & 83.15 & 16.85 \\
(1,234) & \textbf{DRE (ours)} & \textbf{85.58 / 86.36} & \textbf{84.41} &
\textbf{88.32} & \textbf{11.68} \\
\midrule
Cross-source & Objaverse++ & 46.99 / 46.99 & \textbf{88.74} & 5.24 & 94.76 \\
(764) & \textbf{DRE (ours)} & \textbf{88.87 / 88.87} & 80.37 &
\textbf{97.38} & \textbf{2.62} \\
\bottomrule
\end{tabular}
\end{table}

Our method retains 731 of 866 good assets and admits 43 of 368 bad assets.
Objaverse++ retains 245 good assets and admits 62 bad assets. A paired exact
McNemar test gives $p=1.31\times10^{-102}$, with 568 assets correct only for
our method and 63 correct only for Objaverse++. Further we additionally evaluate cross-corpus portability on a
fixed, balanced external reference set of 382 good and 382 bad assets from strict-PBR ABO \cite{collins2022abo} and Objaverse assets. DRE achieves 88.9\% balanced
reference-label accuracy, with 80.4\% good-reference recall and 97.4\%
bad-reference recall. Objaverse got about 47.0\% balanced reference-label
accuracy: although it retains 88.7\% of the strict-PBR reference assets, it
rejects only 5.2\% of the bad assets, admitting 94.8\% as
good.

\begin{figure*}[t]
\centering
\includegraphics[width=\textwidth]{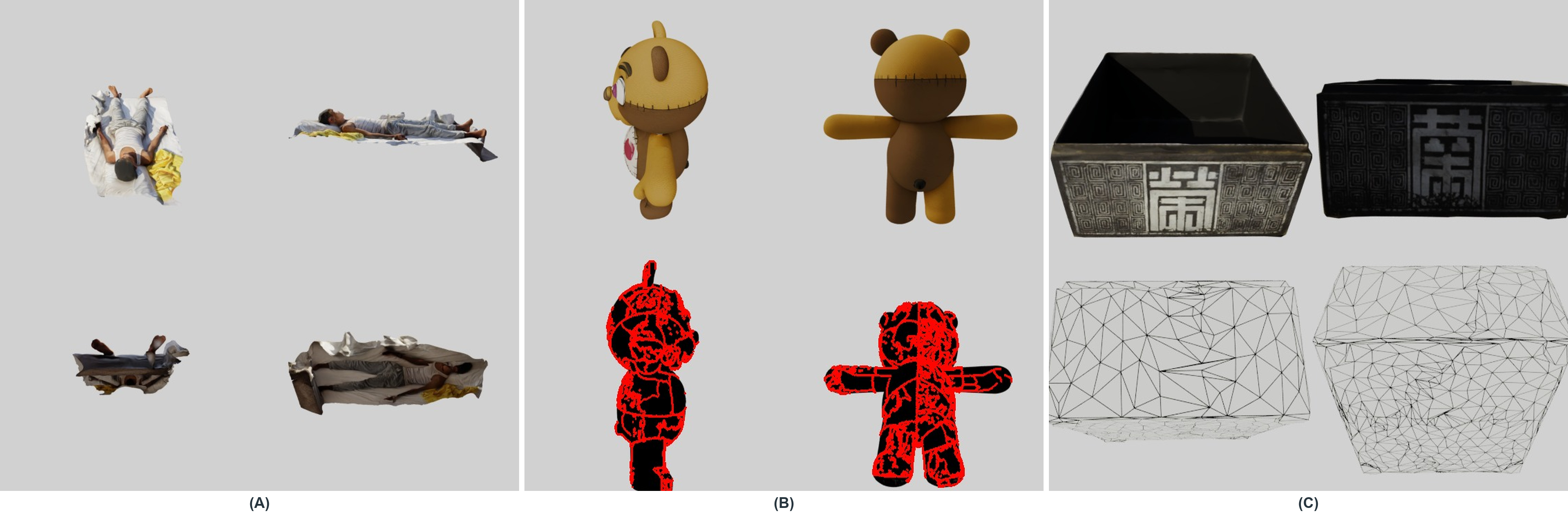}
\caption{Three failure cases revealed by diagnostic rendered evidence. (A) An incomplete scan-like fragment is visibly unusable in beauty views. (B) A plausible-looking character shows unusually broad UV-seam coverage. (C) A regression failure rated \emph{Superior} by the external baseline reveals disconnected and intersecting geometry in wireframe views. These examples illustrate diagnostic evidence rather than universal rejection rules.}
\label{fig:failure-cases}
\end{figure*}

Figure~\ref{fig:failure-cases} shows how diagnostic passes make different failure modes inspectable. Some failures are visible in conventional appearance renders, while others become clear only when the same camera poses expose geometry or UV seams. The diagnostic representation does not turn mesh density or seam count into hard rules. It gives reviewers and the classifier evidence that can be interpreted in the context of the asset and its intended downstream use.

\section{Conclusion}

A beauty render can indicate whether an asset looks plausible, but not whether
it is operationally fit for reliable reuse. Diagnostic evidence for appearance,
topology, UV structure, and material resolution makes the relevant conditions
inspectable from the asset itself. By separating these evidence types before
fusion, DRE provides a repository-agnostic basis for quality-controlled asset
selection for various workflows.

\bibliographystyle{splncs04}
\bibliography{main}

\end{document}